\documentclass{article}

\usepackage[preprint, nonatbib]{nips_2018}

\usepackage[utf8]{inputenc} 
\usepackage[T1]{fontenc}    
\usepackage{hyperref}       
\usepackage{url}            
\usepackage{booktabs, makecell}       
\usepackage{amsfonts}       
\usepackage{nicefrac}       
\usepackage{microtype}      
\usepackage{amsmath}
\usepackage{graphicx}
\usepackage{todonotes}
\usepackage{subcaption}
\usepackage[numbers]{natbib}

\usepackage{algpseudocode,algorithmicx,algorithm}
  
\algrenewcommand\algorithmicrequire{\textbf{Precondition:}}  
\algrenewcommand\algorithmicensure{\textbf{Postcondition:}}

\usepackage{tabularx} 

\title{Stochastic Combinatorial Ensembles for\\ Defending Against Adversarial Examples}

\author{
  George A. Adam  \\
  Department of Computer Science\\
  University of Toronto \\
  Toronto, ON M5S 3G4 \\
  \texttt{alex.adam@mail.utoronto.ca} \\
  \And
  Petr Smirnov \\
  Medical Biophysics \\
  University of Toronto \\
  Toronto, ON M5S 3G4 \\
  \And 
  David Duvenaud \\
  Department of Computer Science \\
  University of Toronto \\
  Toronto, ON M5S 3G4 \\
  \And
  Benjamin Haibe-Kains \\
  Medical Biophysics \\
  University of Toronto \\
  Toronto, ON M5S 3G4
  \And
  Anna Goldenberg \\
  Department of Computer Science \\
  University of Toronto \\
  Toronto, ON M5S 3G4\\
}

\begin{document}

\maketitle

\begin{abstract}
Many deep learning algorithms can be easily fooled with simple adversarial examples.
To address the limitations of existing defenses, we devised a probabilistic framework that can generate an exponentially large ensemble of models from a single model with just a linear cost. This framework takes advantage of neural network depth and stochastically decides whether or not to insert noise removal operators such as VAEs between layers. We show empirically the important role that model gradients have when it comes to determining transferability of adversarial examples, and take advantage of this result to demonstrate that it is possible to train models with limited adversarial attack transferability. Additionally, we propose a detection method based on metric learning in order to detect adversarial examples that have no hope of being cleaned of maliciously engineered noise.
%
%
\end{abstract}

\section{Introduction}

Deep Neural Networks (DNNs) perform impressively well in classic machine learning areas such as image classification, segmentation, speech recognition and language translation \cite{hinton_deep_2012, krizhevsky_imagenet_2012, sutskever_sequence_2014}. These results have lead to DNNs being increasingly deployed in production settings, including self-driving cars, on-the-fly speech translation, and facial recognition for identification. However, like previous machine learning approaches, DNNs have been shown to be vulnerable to adversarial attacks during test time \cite{szegedy_intriguing_2013}. The existence of such adversarial examples suggests that DNNs lack robustness, and might not be learning the higher level concepts we hope they would learn. Increasingly, it seems that the attackers are winning, especially when it comes to white box attacks where access to network architecture and parameters is granted. 

\noindent Several approaches have been proposed to protect against adversarial attacks. Traditional defense mechanisms are designed with the goal of maximizing the perturbation necessary to trick the network, and making it more obvious to a human eye. However, iterative optimization of adversarial examples by computing gradients in a white box environment or estimating gradients using a surrogate model in a black-box setting have been shown to successfully break such defenses.
While these methods are theoretically interesting as they can shed light on the nature of potential adversarial attacks, there are many practical applications in which being perceptible to a human is not a reasonable defense against an adversary. For example, in a self-driving car setting, any deep CNN applied to analyzing data originating from a non-visible light spectrum (e.g. LIDAR), could not be protected even by an attentive human observer. It is necessary to generate `complete' defenses which preclude the existence of adversarial attacks against the model. This requires deeper understanding of the mechanisms which make finding adversarial examples against deep learning models so simple. In this paper, we review characteristics of such mechanisms and propose a novel defense method inspired by our understanding of the problem. In a nutshell, the method makes use of the depth of neural networks to create an exponential population of defenses for the attacker to overcome, and employs randomness to increase the difficulty of successfully finding an attack against the population. 

\section{Related Work}

\noindent Adversarial examples in the context of DNNs have come into the spotlight after Szegedy et al. \cite{szegedy_intriguing_2013}, showed the imperceptibility of the perturbations which could fool state-of-the-art computer vision systems. Since then, adversarial examples have been demonstrated in many other domains, notably including speech recognition \cite{carlini_audio_2018}, and  malware detection\cite{grosse_adversarial_2016}. Nevertheless, Deep Convolutional Neural Networks (CNNs) in computer vision provide a convenient domain to explore adversarial attacks and defenses, both due to the existence of standardized test datasets, high performing CNN models reaching human or super-human accuracy on clean data, and the marked deterioration of their performance when subjected to adversarial examples to which human vision is robust. 


\noindent In order to construct effective defenses against adversarial attacks, it is important to understand their origin. Early work speculated that DNN adversarial examples exist due to the highly non-linear nature of DNN decision boundaries and inadequate regularization, leading to the input space being scattered with small, low probability adversarial regions close to existing data points \cite{szegedy_intriguing_2013}. Follow-up work \cite{goodfellow_explaining_2014} speculated that adversarial examples are transferable due to the linear nature of some neural networks. While this justification did not help to explain adversarial examples in more complex architectures like ResNet 151 on the ImageNet dataset \cite{Liu}, the authors found significant overlap in the misclassification regions of a group of CNNs. In combination with the fact that adversarial examples are transferable between machine learning approaches (for example from SVM to DNN) \cite{papernot_transferability_2016}, this increasingly suggests  that the linearity or non-linearity of DNNs is not the root cause of the existence of adversarial examples that can fool these networks.

\noindent Recent work to systematically characterize the nature of adversarial examples suggests that adversarial subspaces can lie close to the data submanifold \cite{tanay_boundary_2016}, but that they form high-dimensional, contiguous regions of space \cite{tramer_space_2017,ma_characterizing_2018}. This corresponds to empirical observation that the transferability of adversarial examples increases with the allowed perturbation into the adversarial region \cite{carlini_towards_2016}, and is higher for examples which lie in higher dimensional adversarial regions \cite{tramer_space_2017}. Other work \cite{gilmer_adversarial_2018} further suggests that CNNs readily learn to ignore regions of their input space if optimal classification performance can be achieved without taking into account all degrees of freedom in the feature space. In summary, adversarial examples are likely to exist very close to but off the data manifold, in regions of low probability under the training set distribution.

\subsection{Proposed Approaches to Defending Against Adversarial Attacks}

\noindent The two approaches most similar to ours for defending against adversarial attacks are MagNets \citep{meng_magnet:_2017} and MTDeep \citep{sengupta_mtdeep:_2017}. Similar to us, MagNets combine a detector and reformer, based off of the Variational Autoencoder (VAE) operating on the natural image space. The detector is designed by examining the probability divergence of the classification for the original image and the autoencoded reconstruction, with the hypothesis that for adversarial examples, the reconstructions will be classified into the adversarial class with much lower probability. We adapt the probability divergence approach, but instead of relying on the same classification network, we compute the similarity using an explicit metric learning technique, as described in Section \ref{need_for_detection}. MagNets also incorporated randomness into their model by training a population of 8 VAEs acting directly on the input data, with a bias towards differences in their encoding space. At test time, they chose to randomly apply one of the 8 VAEs. They were able to reach 80\% accuracy on defending against a model trained to attack one of the 8 VAEs, but did not evaluate the performance on an attack trained with estimation of the expectation of the gradient over all eight randomly sampled VAEs. We hypothesize that integrating out only 8 discrete random choices would add a trivial amount of computation to mount an attack on the whole MagNet defense. Our method differs in that we train VAEs at each layer of the embedding generated by the classifier network, relying on the differences in embedding spaces to generate diversity in our defenses. This also gives us the opportunity to create a combinatorial growth of possible defenses with the number of layers, preventing the attacker from trivially averaging over all possible defenses. 

\noindent In MTDeep (Moving Target Defense) \cite{sengupta_mtdeep:_2017}, the authors investigate the effect of using a dynamically changing defense against an attacker in the image classification framework. The framework they consider is a defender that has a small number of possible defended classifiers to choose for classifying any one image, and an attacker that can create adversarial examples with high success for each one of the models. They also introduce the concept of differential immunity, which directly quantifies the maximal defense advantage to switching defenses optimally against the attacker. Our method also builds on the idea of constantly moving the target for the attacker to substantially increase the difficulty the attacker has to fool our classifier. However, instead of randomizing only at test time, we use the moving target to make it increasingly costly for the attacker to generate adversarial examples.

\section{Methods}
\subsection{Probabilistic Framework}

\begin{figure}
\includegraphics[width=\textwidth]{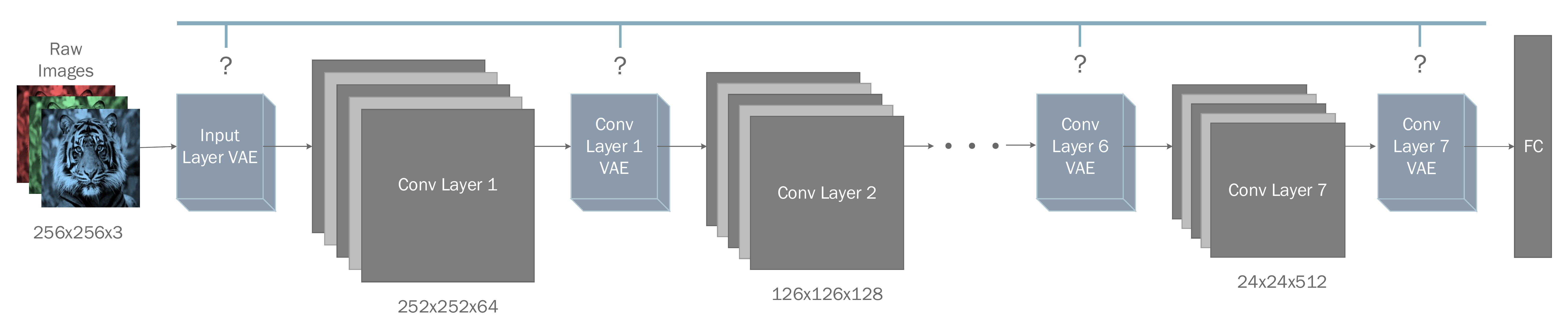}
\caption{Illustration of probabilistic defense framework for a deep CNN architecture. Pooling and activations are left out to save space.}
\label{figure_1}
\end{figure}

Deep neural networks offer many places at which to insert a defense against adversarial attacks. Our goal is to exploit the potential for exponential combinations of defenses. Most defenses based on cleaning inputs tend to operate in image space, prior to the image being processed by the classifier. Attacks that are aware of this preprocessing step can simply create images that fool the defense and classifier jointly. However, an adversary would have increased difficulty attacking a model that is constantly changing. For example, a contracted version of VGG-net has 7 convolutional layers, thus offering $2^{1 + 7} = 64$ (1 VAE before input, and 1 VAE for each of the 7 convolutional layers) different ways to arrange defenses (Figure \ref{figure_1}); this can be thought of as a bag of $64$ models. If the adversary creates an attack for one possible defense arrangement, there is only a $\frac{1}{64}$ chance that the same defense will be used when they try to evaluate the network on the crafted image. Hence, assuming that the attacks are not easily transferable between defense arrangements, the adversary's success rate decreases exponentially as the number of layers increases just linearly. Even if an adversary generates malicious images for all 64 versions of the model, the goal post when testing these images is always moving, so it would take 64 attempts to fool the model on average. An attacker trying to find malicious images fooling all the models together would have to contend with gradient information sampled from random models resulting in possibly orthogonal goals. An obvious defense to use at any given layer is an autoencoder with an information bottleneck such that it is difficult to include adversarial noise in the reconstruction. This should have minimal impact on classifier performance when given normal images, and should be able to clean the noise at various layers in the network when given adversarial images. 



\subsection{Need for Detection} \label{need_for_detection}

Some adversarial examples are generated with large enough perturbations that they no longer resemble the original image, even when judged by humans. It is unreasonable to assume that basic image processing techniques can restore such adversarial examples to their original form since the noise makes the true class ambiguous. Thus, there is a need to flag images with substantial perturbations. Carlini and Wagner \cite{Carlini2017}, demonstrated that methods that try to detect adversarial examples are easily fooled, and also operate under the assumption that adversarial examples are fundamentally different from natural images in terms of image statistics. However, this was in the context of small perturbations. We believe that a good detection method should be effective in detecting adversarial examples that have been generated with large enough perturbations. Furthermore, we believe that detection methods should be generalizable to new types of attacks in order to be of practical relevance, so in our work we do not make any distributional assumptions regarding adversarial noise in building our detector.


%
%

\noindent Our proposed detection method might not seem like a detection method at all. In fact, it is not explicitly trained to differentiate between adversarial examples and natural images. Instead, it relies on the consensus between the predictions of two models: the classifier we are trying to defend, and an auxiliary model. To assure low transferability of adversarial examples between the classifier and auxiliary model, we choose an auxiliary model that is trained in a fundamentally different way than a softmax classifier. This is where we leverage recent developments in metric learning. 

\begin{figure}[h!]\centering

\includegraphics[width=0.8\textwidth]{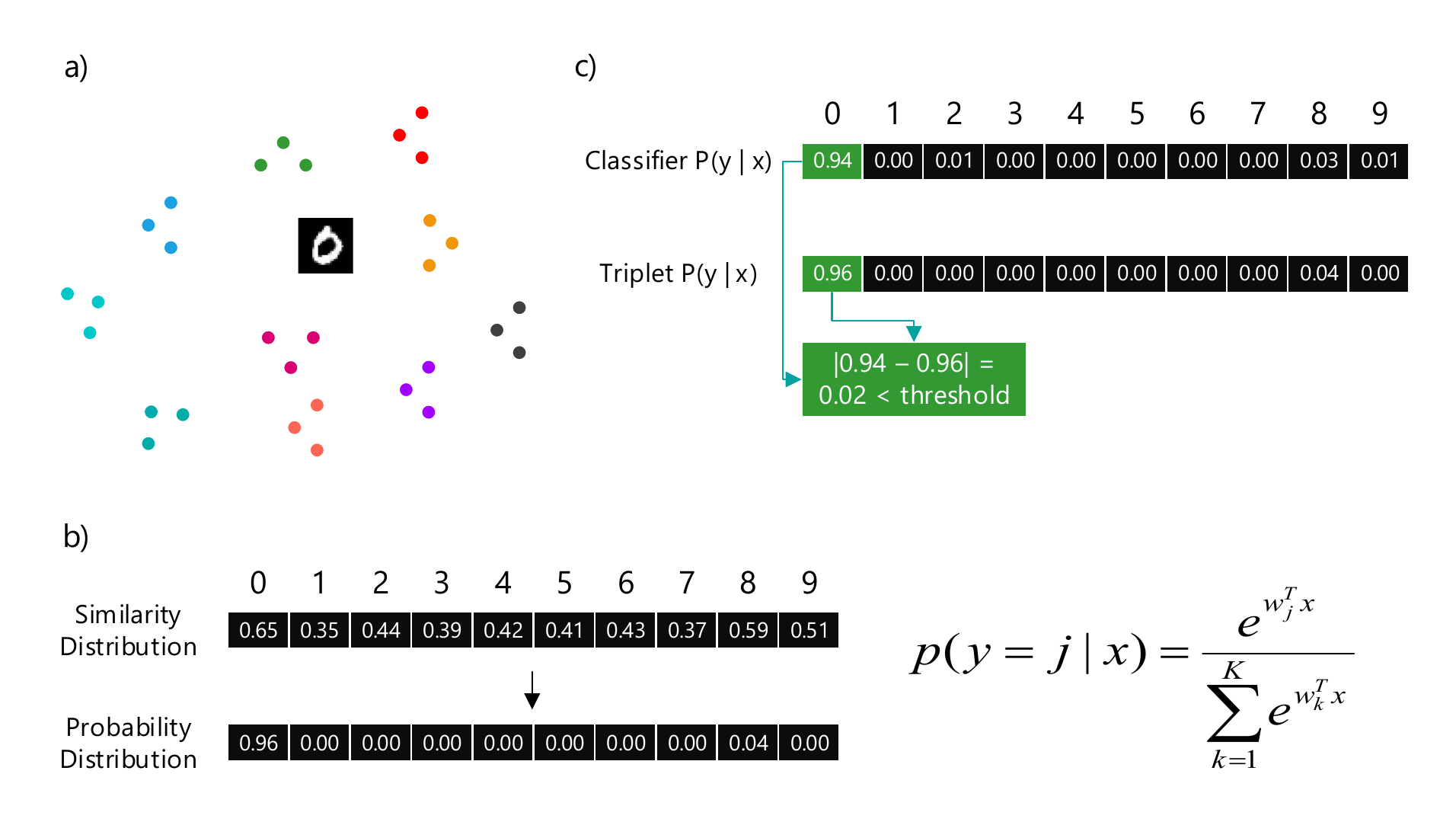}
\caption{a) The test image is projected into embedding space by the embedding network. A fixed number of examples from all possible classes from the training set are randomly sampled. The similarity of the test image embedding to the randomly sampled training embeddings is computed and averaged to get a per-class similarity. b) The vector of similarities is converted to a probability distribution via the softmax function. c) The highest probability class (digit 0 in this case) of the classifier being defended is determined, and the absolute difference between the classifier and triplet probability for that class is computed.}
\label{triplet_detector}
\end{figure}

As an auxiliary model here we use a triplet network \cite{Hoffer}, which was previously introduced in a face recognition application\cite{FaceNet}. A triplet network is trained using 3 different training examples at once: an anchor, a positive example, and a negative example as seen in Supplementary Figure 3; this results in semantic image embeddings that are then used to cluster and compute similarities between images. We use a triplet network for the task of classification via an unconventional type of KNN in the embedding space. This is done by randomly sampling $50$ embeddings for each class from the training set, and then computing the similarity of these embeddings to the embedding for a new test image (Figure \ref{triplet_detector}a). Doing so gives a distribution of similarities that can be then converted to a probability distribution by applying the softmax function (Figure \ref{triplet_detector}b). To classify an image as adversarial or normal, we first take the difference between the probabilities from classifier and from the embedding network (the probabilities are compared for the most likely class of the original classifier). If this difference in probability is high, then the two models do not agree, so the image is classified as adversarial (Figure \ref{triplet_detector}c). Note that for this setup to work, classifiers have to agree in classification of most unperturbed images. In our experiments, we have confirmed the agreement between LeNet and the triplet network is ~90\% as seen in Supplementary Table 2.  More formally, the detector uses the following logic

\begin{align*}
k &= \mathrm{argmax}(p_c(y | x)) & \Delta &= |p_c(y = k | x) - p_t(y = k | x)| & D(\Delta) = 
	\begin{cases} 
      1 & \Delta \geq \eta \\
      0 & \mathrm{otherwise} 
   \end{cases}
\end{align*}

\noindent where $p_c(y | x)$ and $p_t(y | x)$ are the probability distributions from the classifier and triplet network respectively, $k$ is the most probable class output by the classifier, and $\Delta$ is the difference in probability between the most probable class output by the classifier and the same class output by the triplet network. In our experiments we set the threshold $\eta=0.4$. Note that while we used a triple network as an auxiliary model in our examples, the goal is to find a model that is trained in a distinct manner and will thus have different biases than the original classifier, so other models can certainly be used in place of a triplet network if desired. 

\section{Defense Analysis}

Before discussing the experiments and corresponding results, here is the definition of adversarial examples that we are working with in this paper. An image is an adversarial example if 
\begin{enumerate}
\item It is classified as a different class than the image from which it was derived.
\item It is classified incorrectly according to ground truth label.
\item The image from which it was derived was originally correctly classified.
\end{enumerate}

Since we are evaluating attack success rates in the presence of defenses, point 3 in the above definition ensures that the attack success rate is not confounded by a performance decrease in the classifier potentially caused by a given defense. In our analysis, we use the fast gradient sign (FGS) \cite{goodfellow_explaining_2014}, iterative gradient sign (IGS) \cite{kurakin_adversarial_2016}, and Carlini and Wagner (CW) \cite{carlini_towards_2016} attacks. We are operating under the white-box assumption that the adversary has access to network architecture, weights, and gradients.

Additionally, we did not use any data augmentation when training our models since that can be considered as a defense and could further confound results.
To address possible concerns regarding poor performance on unperturbed images when defenses are used, we performed the experiment below. For the FGS and IGS attacks, unless otherwise noted, an epsilon of 0.3 was used as is typical in the literature.

\subsection{Performance of Defended Models on Clean Data}

One of the basic assumptions of our approach is that there exist operations that can be applied to the feature embeddings generated by the layers of a deep classification model, which preserve the classification accuracy of the network while removing the adversarial signal. As an example of such transformations, we propose a variational autoencoder.
We have evaluated the effect of inserting  VAEs on two models: Logistic Regression (LR) and a 2 convolutional layer LeNet on the MNIST dataset. The comparison of the performance of these methods is summarized in Table \ref{table:performance_reduction}. Surprisingly, on MNIST it is possible to train quite simple variational autoencoding models to recreate feature embeddings with sufficient fidelity to leave the model performance virtually unchanged. Reconstructed embeddings are visualized in Supplementary Materials. Supplementary Figure 3 shows how the defense reduces the distance between adversarial and normal examples in the various layers of LeNet.

\begin{table}[!h]
\caption{Performance reduction caused by defenses on the MNIST dataset.}
\label{table:performance_reduction}
\centering
\begin{tabular}{lccc}  
\toprule
Model & Undef. Accuracy & Deterministic Def. Accuracy & Stochastic Def. Accuracy \\
\midrule
LR-VAE      & 0.921    & 0.907  & 0.914   \\
LeNet-VAE   & 0.990    & 0.957  & 0.972   \\
\bottomrule
\end{tabular}
\end{table}

\subsection{Transferability of Attacks Between Defense Arrangements}

The premise of our defense is that the exponentially many arrangements of noise removing operations are not all exploitable by the same set of adversarial images. The worst case scenario is if the adversary creates malicious examples when noise removing operations are turned on in all possible locations. It is possible that such adversarial examples would also fool the classifier when the defense is only applied in a subset of the layers. Fortunately, we note that for FGS, IGS, and CW2, transferability of attacks between defense arrangements is limited as seen in Table \ref{table:transferability_layer_lenet}.
The column headers are binary strings indicating presence or absence of defense at the 3 relevant points in LeNet: input layer, after the first convolutional layer, after the second convolutional layer. Column [0, 0, 0] shows the attack success rate against an undefended model, and column [1, 1, 1] shows the attack success rate against a fully deterministically-defended model. The remaining columns show the transfer attack success rate of the perturbed images created for the [1, 1 ,1] defense arrangement. The most surprising result is that defense arrangements using 2 autoencoders are more susceptible to a transfer attack than defense arrangements using a single autoencoder. Specifically, [1, 0, 0] is more robust than [1, 1, 0] which does not make sense intuitively. Although the perturbed images are engineered to fool a classifier with VAEs in all convolutional layers in addition to the input layer, it is possible that the gradient used to generate such images is orthogonal to the gradient required to fool just a single VAE in the convolutional layers.

\begin{table}
\caption{Transferability of attacks from strongest defense arrangement [1, 1, 1] to other defense arrangements for LeNet-VAE.}
\label{table:transferability_layer_lenet}
\centering
\begin{tabular}{lcccccccc}  
\toprule
\multicolumn{1}{c}{} & \multicolumn{2}{c}{Success Rate} & \multicolumn{6}{c}{Transfer Success Rate} \\
\cmidrule(r){2-3} \cmidrule(r){4-9}
 Attack & [0, 0, 0] & [1, 1, 1] & [0, 0, 1] & [0, 1, 0] & [1, 0, 0] & [0, 1, 1] & [1, 0, 1] & [1, 1, 0]\\
\midrule
 FGS & 0.176 & 0.1451 & 0.035 & 0.019 & 0.117 & 0.018 & 0.133 & 0.131	 \\
 IGS & 0.434 & 0.270 & 0.016 & 0.011 & 0.193 & 0.014 & 0.231 & 0.223 \\
 CW2 & 0.990 & 0.977 & 0.003 & 0.002 & 0.775 & 0.003 & 0.959 & 0.892 \\
\bottomrule
\end{tabular}
\end{table}

\subsection{Investigating Cause of Low Attack Transferability}

To confirm the suspicion that orthogonal gradients are responsible for the unexpected transferability results between defense arrangements seen in Table \ref{table:transferability_layer_lenet}, we computed the cosine similarity of the gradients of the output layer w.r.t to the input images. Table \ref{table:cosine_similarities_lenet} shows the average cosine similarity between the strongest defense arrangement [1, 1, 1] and other defense arrangements. To summarize the relationship between cosine similarity and attack transferability, we computed the correlations of the transferabilities in Table \ref{table:transferability_layer_lenet} with the cosine similarities in Table \ref{table:cosine_similarities_lenet}. These correlations are shown in Table \ref{table:pearson_correlations}. It is quite clear that cosine similarity between gradients is an almost perfect predictor of the transferability between defense arrangements. Thus, training VAEs with the goal of gradient orthogonality, or training conventional ensembles of models with this goal has the potential to drastically decrease the transferability of adversarial examples between models. 


\begin{table}[h!]
\caption{Cosine similarities of LeNet probability output vector gradients w.r.t. input images.}
\label{table:cosine_similarities_lenet}
\centering
\begin{tabular}{lccccccc}  
\toprule
\multicolumn{1}{c}{} & \multicolumn{7}{c}{Transfer Arrangement Cosine Similarity} \\
\cmidrule(r){2-8}
 Base Arrangement & [0, 0, 1] & [0, 1, 0] & [0, 1, 1] & [1, 0, 0] & [1, 0, 1] & [1, 1, 0] & [1, 1, 1] \\
\midrule
 {[1, 1, 1]} & 0.219 & \textbf{0.190} & 0.249 & 0.648 & 0.773 & 0.728  & 0.949 \\
\bottomrule
\end{tabular}
\end{table}

\begin{table}[h!]
\caption{Correlations of cosine similarities between different defense arrangements (using [1, 1, 1] as the baseline defense) and the transferability between the attacks on the [1, 1, 1] defense to other defense arrangements.}
\label{table:pearson_correlations}
\centering
\begin{tabular}{lccc}
\toprule
\multicolumn{1}{c}{} & \multicolumn{3}{c}{Correlation} \\
\cmidrule(r){2-4}
Cor Type & FGS & IGS & CW2 \\
\midrule
Pearson & 0.990 & 0.997 & 0.997 \\
Spearman & 0.829 & 0.943 & 0.986 \\
\bottomrule
\end{tabular}
\end{table}

\subsection{Training for Gradient Orthogonality}

\begin{figure}[H]
\centering
\includegraphics[width=0.9\textwidth]{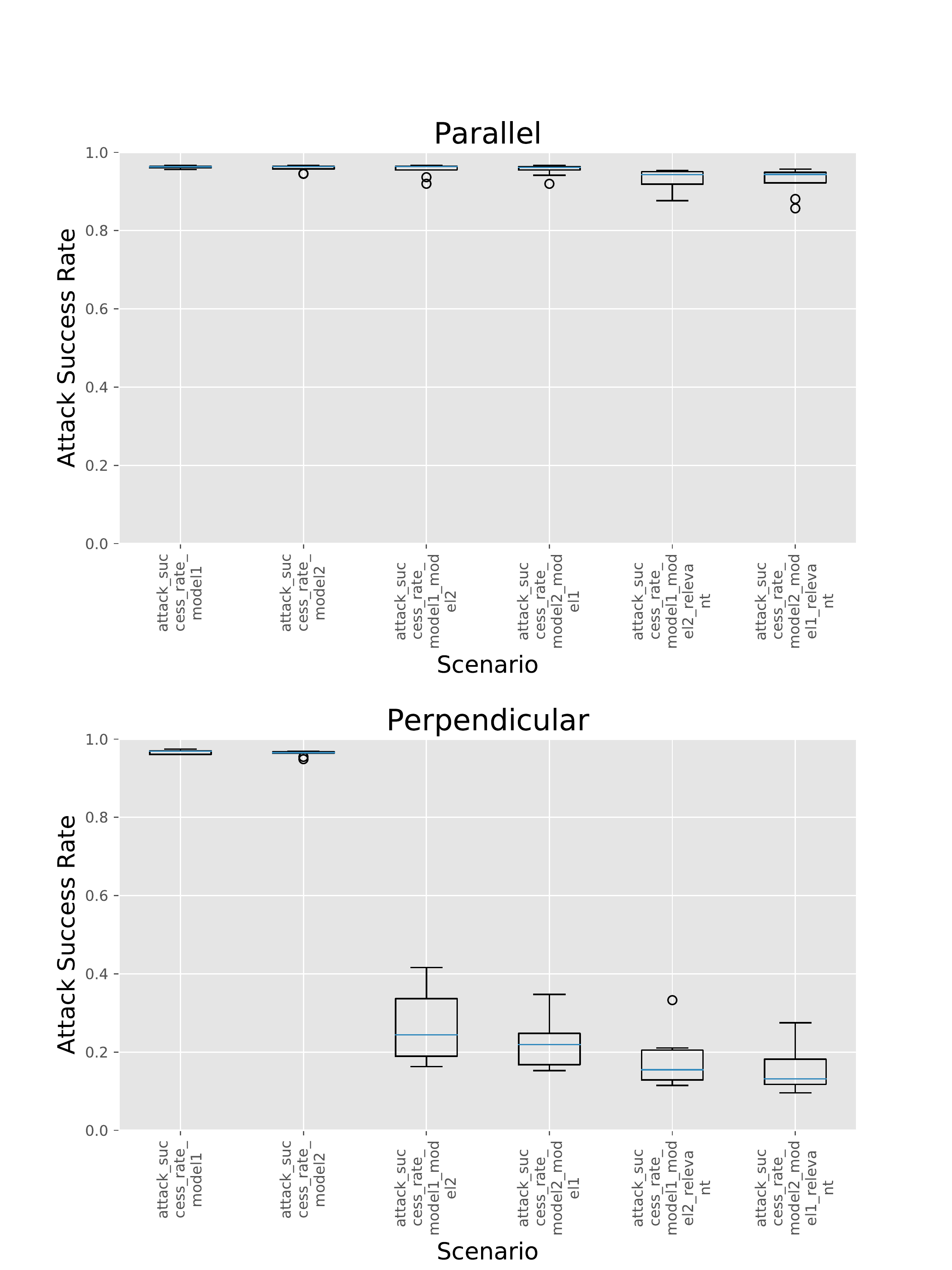}
\caption{Baseline attack success rates and transfer success rates for an IGS attack with an epsilon of 1.0 on LeNet models trained on MNIST. 8 pairs of models were trained for the parallel Jacobian goal, and 8 pairs of models were trained for the perpendicular goal to obtain error bars around attack success rates.}
\label{fig:parallel_vs_perpendicular}
\end{figure}

In order to test the claim that explicitly training for gradient orthogonality will result in lower transferability of adversarial examples, we focus on a simple scenario. We trained 16 pairs of LeNet models, 8 of which were trained to have orthogonal input-output Jacobians, and 8 of which were trained to have parallel input-output Jacobians. As can be seen in Figure \ref{fig:parallel_vs_perpendicular}, the differences in transfer rates and relevant transfer rates are quite vast between the two approaches. The median relevant transfer attack success rate for the parallel approach is approx. 92\%, whereas it is only approx. 17\% for the perpendicular approach. This result further illustrates the importance of the input-output Jacobian cosine similarity between models when it comes to transferability.


\subsection{Effect of Gradient Magnitude}

\begin{figure}[H]
\includegraphics[width=\textwidth]{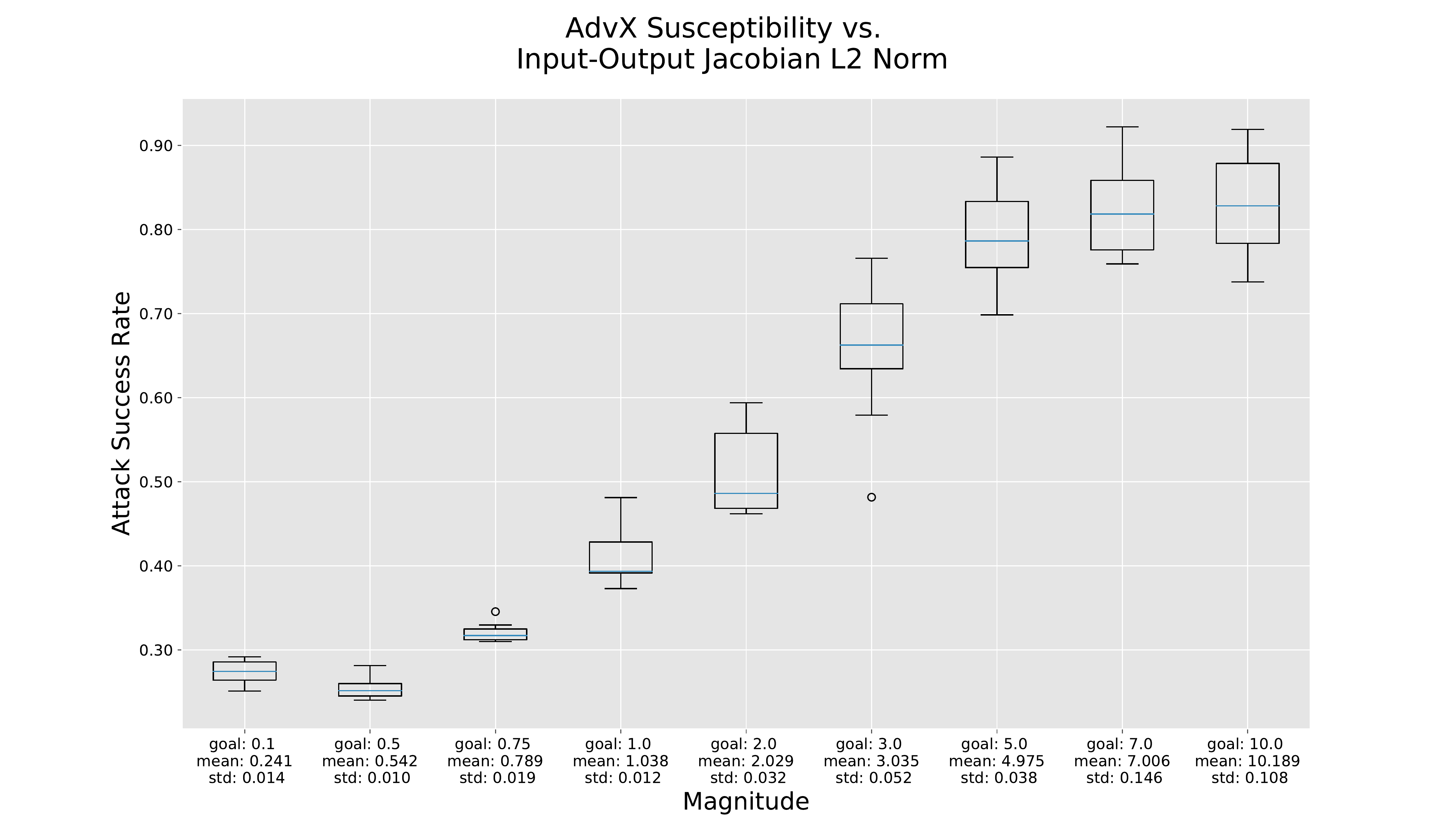}
\caption{Relationship between input-output Jacobian L2 norm and susceptibility to IGS attack with epsilon of 0.3. 8 LeNet models were trained on MNIST for each magnitude with different random seeds. The 'mean' and 'std' reported underneath the goal note the actual input-output Jacobian L2 norms, whereas the 'goal' is the target that was used during training to regularize the magnitude of the Jacobian.}
\label{fig:effect_of_grad_magnitude}
\end{figure}

When training the dozens of models used for the results of this paper, we noticed that attack success rates varied greatly between models, even though the difference in hyperparameters seemed negligible. We posited that a significant confounding factor that determines susceptibility to attacks such as FGS and IGS is the magnitude of the input-output Jacobian for the true class. Intuitively, if the magnitude of the input-output Jacobian is large, then the perturbation required to cause a model to misclassify an image is smaller than had the magnitude of the input-output Jacobian been small. This is seen in Figure \ref{fig:effect_of_grad_magnitude} where there is a clear increasing trend in attack success rate as the input-output Jacobian increases. This metric can be a significant confounding factor when analyzing robustness to adversarial examples, so it is important to measure it before concluding that differences in hyperparameters are the cause of varying levels of adversarial robustness. 

\section{Detector Analysis}
The effect of the CW2 attack on the average of defense arrangements is investigated in Supplementary Material. While it may seem that our proposed defense scheme is easily fooled by a strong attack such as CW2, there are still ways of recovering from such attacks by using detection. In fact, there will always be perturbations that are extreme enough to fool any classifier, but perturbations with larger magnitude become increasingly easy for humans to notice. In this section, "classifier" is the model we are trying to defend, and "auxiliary model" is another model we train (a triplet network) that is combined with the classifier to create a detector.

\subsection{Transferability of Attacks Between LeNet and Triplet Network}

The best case scenario for an auxiliary model would be if it were fooled only by a negligible percentage of the images that fool the classifier. It is also acceptable for the auxiliary model to be fooled by a large percentage of those images, provided it does not make the same misclassifications as the classifier. Fortunately, we observe that the majority of the perturbed images that fooled LeNet did not fool the triplet network in a way that would affect the detector's viability. This is shown in Table \ref{table:transferability_classifier_detector}. For example, 1060 perturbed images created for LeNet using FGS fooled the triplet network. However, only 70 images were missed by the detector due to the requirement for agreement between the auxiliary model and classifier. The columns with "Relevant" in the name show the success rate of transfer attacks that would have fooled the detector. 

\begin{table}[h!]
\caption{Transferability of attacks between LeNet and triplet network.}
\label{table:transferability_classifier_detector}
\centering
\begin{tabular}{lcccccc}
\toprule
\multicolumn{1}{c}{} & \multicolumn{6}{c}{Attack Success Rate} \\
\cmidrule(r){2-7}
\thead{Attack} & \thead{LeNet} & \thead{LeNet to \\Triplet} & \thead{Triplet} & \thead{Triplet to \\ LeNet} & \thead{Relevant \\ LeNet to Triplet} & \thead{Relevant \\ Triplet to LeNet} \\
\midrule
 FGS & 0.089 & 0.106 & 0.164 & 0.015 & 0.007 & 0.004 \\
 IGS & 0.130 & 0.094 & 0.244 & 0.004 & 0.007 & 0.002\\
 CW2 & 0.990 & 0.121 & 0.819 & 0.013 & 0.049 & 0.008 \\
\bottomrule
\end{tabular}
\end{table}

\subsection{Jointly Fooling Classifier and Detector}

If an adversary is unaware that a detector is in place, the task of detecting adversarial examples is much easier. To stay consistent with the white-box scenario considered in previous sections, we assume that the adversary is aware that a detector is in place, so they choose to jointly optimize fooling the VAE-defended classifier and detector. We follow the approach described in  \cite{Carlini2017} where we add an additional output as follows

\[
  G(x)_i =
  \begin{cases}
                                   Z_F(x)_i & \text{if $i \leq N$} \\
                                   (Z_D(x) + 1) \cdot \max\limits_{j} Z_F(x)_j & \text{if $i=N + 1$}
  \end{cases}
\]

where $Z_D(x)$ is the logit output by the detector, and $Z_F(x)$ is the logit output by the classifier. Table \ref{table:lenet_detector} shows the effectiveness of combining the detector and VAE-defended classifier. The reason why this undefended attack success rate for the CW2 attack is lower than that in Table \ref{table:transferability_classifier_detector} is probably because the gradient signal when attacking the joint model is weaker. Overall, less than 7\% (0.702 - 0.635) of the perturbed images created using CW2 fool the combination of VAE defense and detector.

\begin{table}[h!]
\caption{Attack success rates and detector accuracy for adversarial examples on LeNet-VAE using a triplet network detector.}
\label{table:lenet_detector}
\centering
\begin{tabular}{lccccccc}
\toprule
\multicolumn{1}{c}{} & \multicolumn{3}{c}{Attack Success Rate} & \multicolumn{4}{c}{Detector Accuracy} \\
\cmidrule(r){2-4} \cmidrule(r){5-8}
Attack & Undef. & Determ. & Stoc. & Original & Undefended & Deterministic & Stochastic \\
 FGS & 0.197 & 0.178 & 0.179 & 0.906 & 0.941 & 0.957 & 0.962 \\
 IGS & 0.323 & 0.265 & 0.146 & 0.903 & 0.938 & 0.949 & 0.967\\
 CW2 & 0.787 & 0.703 & 0.702 & 0.909 & 0.848 & 0.635 & 0.657 \\
\bottomrule
\end{tabular}
\end{table}

\section{Discussion}

Our proposed defense and the experiments we conducted on it have revealed intriguing, unexpected properties of neural networks. Firstly, we showed how to obtain an exponentially large ensemble of models by training a linear number of VAEs. Additionally, various VAE arrangements had substantially different effects on the network's gradients w.r.t. a given input image. This is a counterintuitive result because qualitative examination of reconstructions shows that reconstructed images or activation maps look nearly identical to the original images. Also, from a theoretical perspective, VAE reconstructions in general should have a gradient of $\approx \mathbf{1}$ w.r.t. input images since VAEs are trained to approximate the identity function. Secondly, we demonstrated that reducing the transferability of adversarial examples between models is a matter of making the gradients orthogonal between models w.r.t. the inputs. This result makes more sense, and such a goal is something that can be enforced via a regularization term when training VAEs or generating new filtering operations. Using this result can also help guide the creation of more effective concordance-based detectors. For example, a detector could be made by training an additional LeNet model with the goal of making the second model have the same predictions as the first one while having orthogonal gradients. Conducting an adversarial attack that fools both models in the same way would be difficult since making progress on  the first model might have the opposite effect or no effect at all on the second model. 

\noindent 
A limitation of training models with such unconventional regularization in place is determining how to trade off classification accuracy and gradient orthogonality. Our defense framework requires little computational overhead to filter operations such as blurs and sharpens, and is not particularly computationally intensive when there are VAEs to train. Training a number of VAEs equal to the depth of a network in order to obtain an ensemble containing an exponentially large number of models can be computationally intensive, however, in critical mission scenarios, such as healthcare and autonomous driving, spending more time to train a robust system is certainly warranted and is a key to broad adoption. 



\section{Conclusion}

In this project, we presented a probabilistic framework that uses properties intrinsic to deep CNNs in order to defend against adversarial examples. Several experiments were performed to test the claims that such a setup would result in an exponential ensemble of models for just a linear computation cost. We  demonstrated that our defense cleans the adversarial noise in the perturbed images and makes them more similar to natural images (Supplementary). Perhaps our most exciting result is that the cosine similarity of the gradients between defense arrangements is highly predictive of attack transferability which opens a lot of avenues for developing defense mechanisms of CNNs and DNNs in general.  As proof of a concept regarding classification biases between models, we showed that the triplet network detector was quite effective at detecting adversarial examples, and was fooled by only a small fraction of the adversarial examples that fooled LeNet. To conclude, probabilistic defenses are able to substantially reduce adversarial attack success rates, while revealing interesting properties about existing models.


\bibliographystyle{unsrtnat}
\bibliography{sample,Zotero}

\newpage

\section*{Supplementary Material}
\subsection*{Reconstruction of Feature Embeddings}
\label{appdx:reconstruct}

\begin{figure}[H]
\centering
\includegraphics[width=5cm,height=15cm,keepaspectratio]{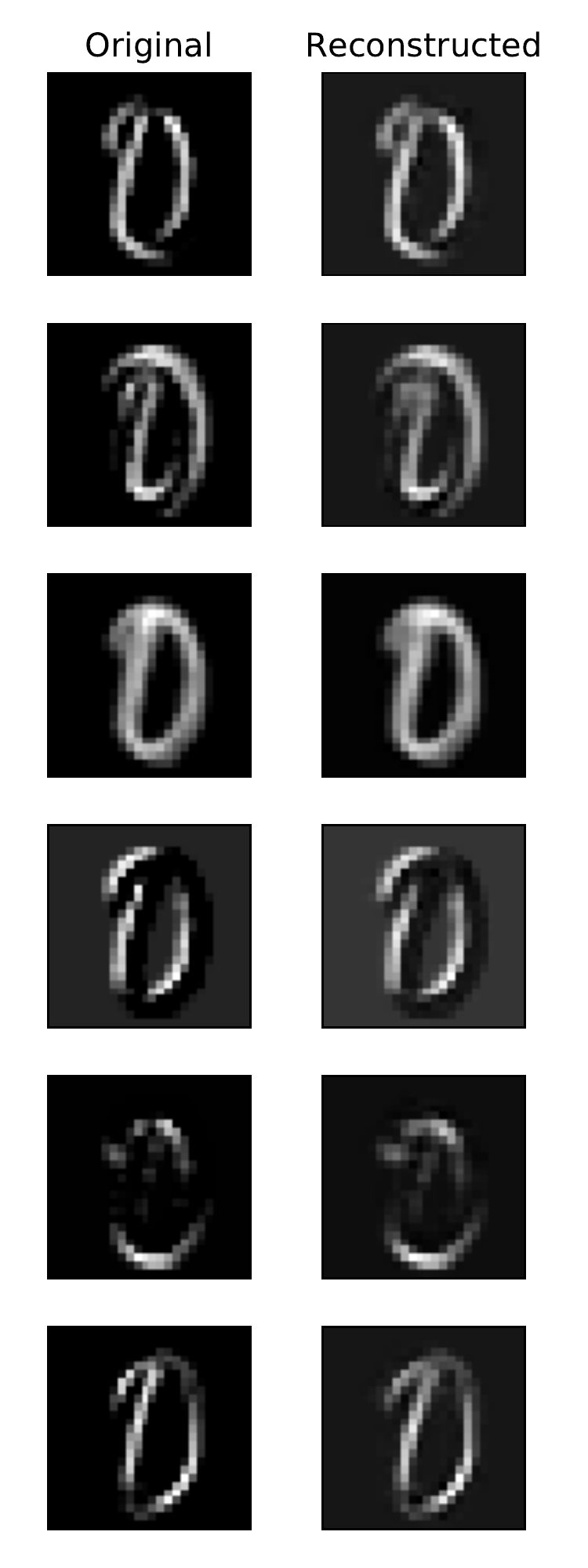} 
\caption{First convolutional layer feature map visualization for LeNet on MNIST. Original feature maps are on the left, VAE reconstructed features maps are on the right. As is seen, reconstructions are of very high quality.}
\label{fig:conv1_layer_reconstructions}
\end{figure}

\begin{figure}[H]
\centering
\includegraphics[width=5cm,height=15cm,keepaspectratio]{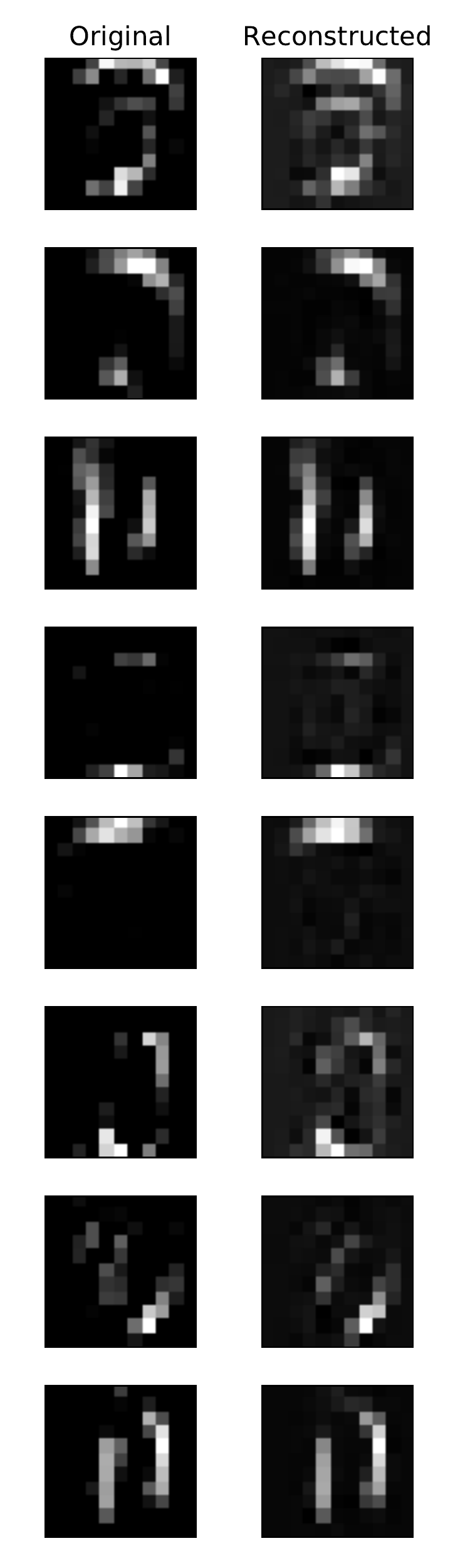} 
\caption{Second convolutional layer feature map visualization for LeNet on MNIST. Original feature maps are on the left, VAE reconstructed features maps are on the right. As is seen, reconstructions are of very high quality.}
\label{fig:conv2_layer_reconstructions}
\end{figure}

Since using autoencoders to reconstruct activation maps is an unconventional task, we visualize the reconstructions in order to inspect their quality. For the activations obtained from the first convolutional layer seen in Figure \ref{fig:conv1_layer_reconstructions}, it is obvious that the VAEs are effective at reconstructing the activation maps. The only potential issue is that the background for some of the reconstructions is slightly more gray than in the original activation maps. For the most part, this is also the case for the second convolutional layer activation maps. However, in the first, fourth, and sixth rows of Figure \ref{fig:conv2_layer_reconstructions}, there is an obvious addition of arbitrary pixels that were not present in the original activation maps.

\subsection*{Cleaning Adversarial Examples}

\begin{figure}[h!]
\centering
\begin{subfigure}{.5\textwidth}
  \centering
  \includegraphics[width=1\linewidth]{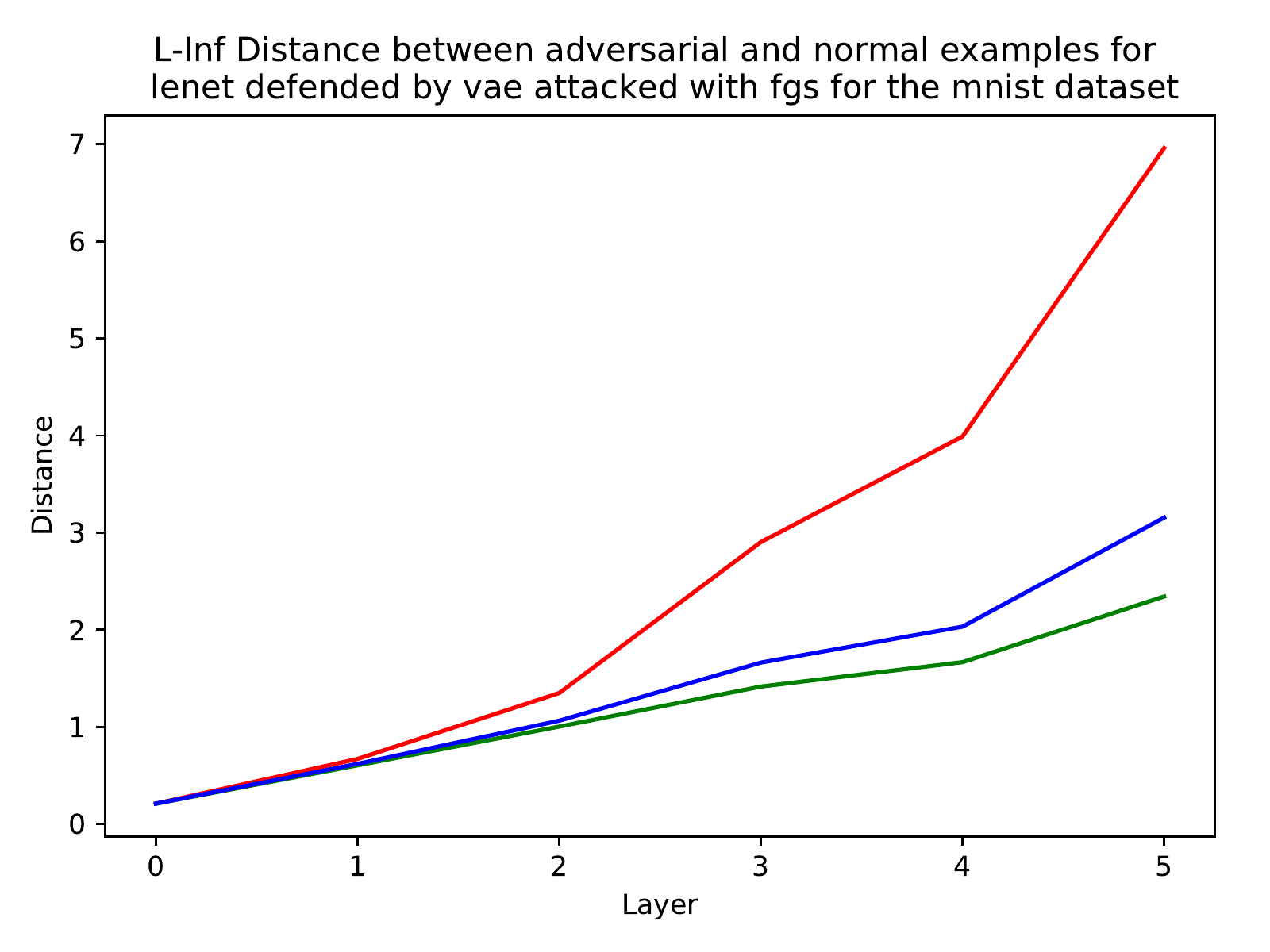}
  \caption{VAE Defence}
  \label{fig:vae}
\end{subfigure}
\caption{L-$\infty$ distance between adversarial and normal images as a function of layer number for LeNet attacked with FGS for the MNIST dataset.}
\label{fig:lenet_fgs_distances}
\end{figure}

The intuitive notion that VAEs or filters  remove adversarial noise can be tested empirically by comparing the distance between adversarial examples and their unperturbed counterparts. In figure \ref{fig:lenet_fgs_distances}, the evolution of distances between normal an adversarial examples can be seen. When the classifier is undefended, the distance increases significantly with the depth of the network, and this confirms the hypothesis that affine transformations amplify noise. However, it is clear that applying our defense has a marked impact on the distance between normal and adversarial examples. Thus, we can conclude that part of the reason for why the defense works is that it dampens the effect of adversarial noise.

\subsection*{Effect of Attacks on Averaged Defense}

\begin{figure}[h!]
\centering
\includegraphics[width=0.8\linewidth]{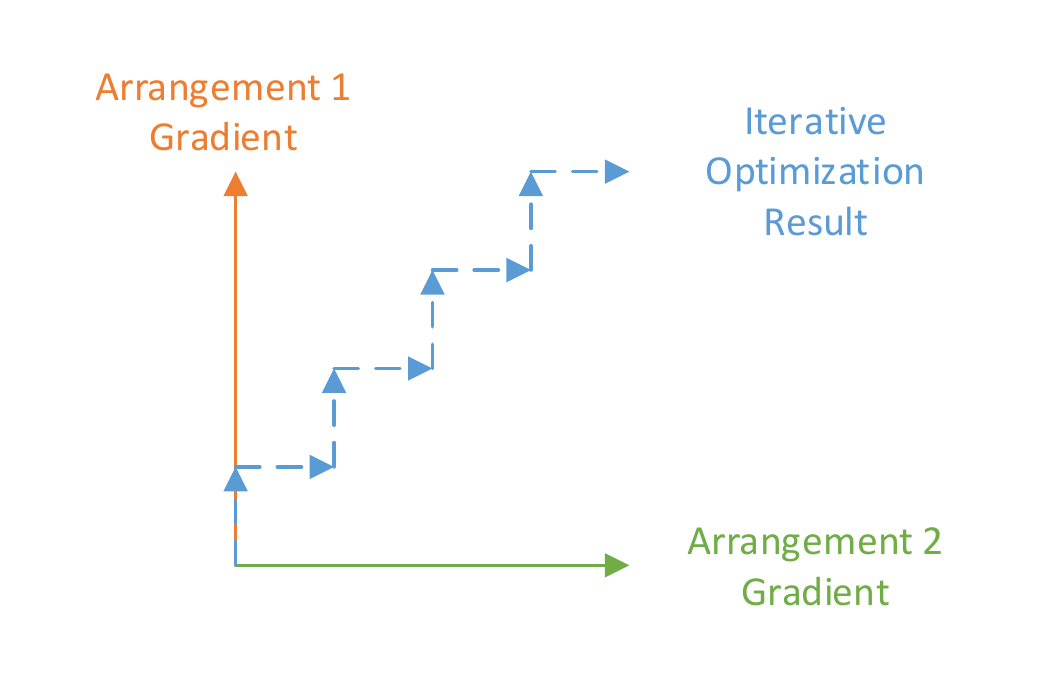}
\caption{Illustration of how the defense can fail against iterative attacks. Even though the two defense arrangements have orthogonal gradients, thereby exhibiting low transferability of attacks, an iterative attack that alternates between optimizing for either arrangement can end up fooling both.}
\label{fig:failure_mode}
\end{figure}

Since the IGS and CW2 attacks are iterative, they have the ability to see multiple defense arrangements while creating adversarial examples. This can result in adversarial examples that might fool any defense of the available defense arrangements. Indeed, this seems to happen for the CW2 attack shown in Table \ref{table:defense_success_lenet}. The cause of this is most easily explained by the illustration in Figure \ref{fig:failure_mode}. Since the depth of the models we trained was not deep enough, it was possible for the iterative attacks to see all defense combinations when creating adversarial examples, so our defense was defeated. We believe that given a deep enough network of 25 or more layers, it would be computationally infeasible for an adversary to create examples that fool the stochastic ensemble.

\begin{table}[h!]
\caption{Success rate of CW2 attack on LR and LeNet defended with VAEs. }
\label{table:defense_success_lenet}
\centering
\begin{tabular}{lccccc}
 \toprule
 \thead{Model} & \thead{Undefended} & \thead{Deterministic \\ Defence}  & \thead{Deterministic \\ Defence \\ Reattacked} & \thead{Stochastic \\ Defence} & \thead{Stochastic \\ Defence \\ Reattacked} \\
 \midrule
 LR-VAE & 0.920 & 0.032 & 0.922 & 0.473 & 0.921 \\
 LeNet-VAE & 0.990 & 0.014 & 0.977 & 0.140 & 0.984 \\
 \bottomrule
\end{tabular}

\end{table}

\subsection*{Triplet Network Visualization}

Here we illustrate how a triplet network is trained. An anchor, positive example, and negative example are all passed through the same embedding network. The triplet loss is then computed which encourages the distance between the anchor and positive example to be some margin $\alpha$ closer together than the anchor and negative example.

\begin{figure}[H]
\includegraphics[width=\textwidth]{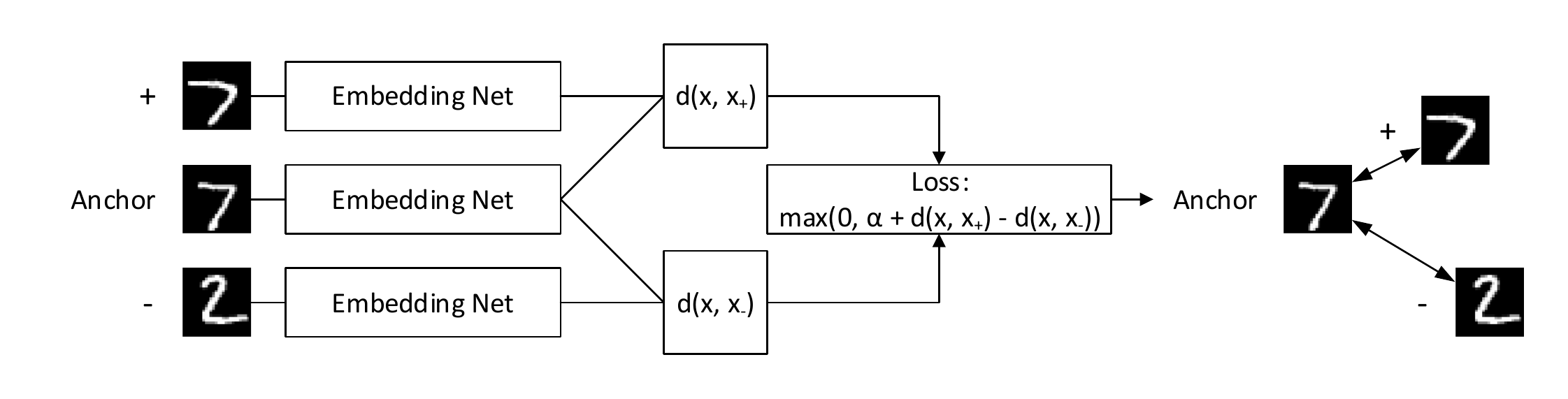}
\caption{Illustration of how a triplet network works on the MNIST dataset.}
\label{triplet_net}
\end{figure}

\noindent The triplet loss function is shown here for convenience:

\begin{equation}
L(x, x_{-}, x_{+}) = \mathrm{max}(0, \alpha + \mathrm{d}(x, x_{+}) - \mathrm{d}(x, x_{-}))
\end{equation}

\subsection*{Agreement Between LeNet and Triplet Network}

We investigated the agreement between LeNet and triplet network we trained in order to confirm that a concordance based detector is a viable option, and does not result in false positives on normal images. Importantly, the models agree ~90\%  (Table \ref{table:concordance}) of the time on normal images, so false positives are not a major concern.

\begin{table}[h!]
\caption{Concordance between LeNet and triplet network on predictions on normal images.}
\label{table:concordance}
\centering
\begin{tabular}{ccc}
\toprule
Overall Concordance & Correct Concordance & Incorrect Concordance \\
\midrule
0.911 & 0.908 & 0.003
\end{tabular}
\end{table}

\end{document}